\pdfoutput=1

\documentclass[11pt]{article}

 \usepackage[final]{acl}

\usepackage{times}
\usepackage{latexsym}

\usepackage[T1]{fontenc}

\usepackage[utf8]{inputenc}

\usepackage{microtype}

\usepackage{inconsolata}

\usepackage{graphicx}
\usepackage{cleveref}
\usepackage{subcaption}
\usepackage[underline=false]{pgf-umlsd}

\usepackage{todonotes}
\usepackage{xspace}

\renewenvironment{call}[5][1]{
\ifthenelse{\equal{#2}{#4}}
{
  \begin{callself}[#1]{#2}{#3}{#5}
}
{
  \begin{callanother}[#1]{#2}{#3}{#4}{#5}
}
}
{
\ifthenelse{\equal{\f\thecallevel}{\t\thecallevel}}
{
  \end{callself}
}
{
  \end{callanother}
}
}

\renewenvironment*{callanother}[7][1]{
  \stepcounter{seqlevel}
  \stepcounter{callevel} 
  \path
  (#2)+(0,-\theseqlevel*\unitfactor-0.7*\unitfactor) node (cf\thecallevel) {}
  (#4.\threadbias)+(0,-\theseqlevel*\unitfactor-0.7*\unitfactor) node (ct\thecallevel) {};

  \def\l\thecallevel{#1}
  \def\f\thecallevel{#2}
  \def\t\thecallevel{#4}
  \def\returnvalue{#5}
  \def\calloptions{#6}
  \def\returnoptions{#7}

  \ifx\calloptions\empty
    \def\calloptions{midway, fill=white, inner sep=0, inner xsep=1mm}%
  \fi
  \ifx\returnoptions\empty
    \def\returnoptions{midway, fill=white, inner sep=0, inner xsep=1mm}%
  \fi

  \begingroup\edef\x{\endgroup
    \noexpand\draw[->,>=triangle 60] ({cf\thecallevel}) -- (ct\thecallevel) node[\calloptions] {#3};
  }\x
  \tikzstyle{threadstyle}+=[instcolor#2]
}
{
  \addtocounter{seqlevel}{\l\thecallevel}
  \path
  (\f\thecallevel)+(0,-\theseqlevel*\unitfactor-0.7*\unitfactor) node (rf\thecallevel) {}
  (\t\thecallevel.\threadbias)+(0,-\theseqlevel*\unitfactor-0.7*\unitfactor) node (rt\thecallevel) {};
  \begingroup\edef\x{\endgroup
    \noexpand\draw[dashed,->,>=angle 60] ({rt\thecallevel}) -- (rf\thecallevel)
    node[\returnoptions]{\noexpand\returnvalue};
  }\x
  \drawthread{ct\thecallevel}{rt\thecallevel}
  \addtocounter{callevel}{-1} 
}

\title{A Browser-based Open Source Assistant for  Multimodal Content  Verification}

\author{
 \textbf{Rosanna Milner\textsuperscript{1}},
 \textbf{Michael Foster\textsuperscript{1}},
 \textbf{Twin Karmakharm\textsuperscript{1}}
 \textbf{Olesya Razuvayevskaya\textsuperscript{1}},
 \\
 \textbf{Ian Roberts\textsuperscript{1}},
 \textbf{Valentin Porcellini\textsuperscript{2}},
 \textbf{Denis Teyssou\textsuperscript{2}},
 \textbf{Kalina Bontcheva\textsuperscript{1}},
\\
\\
 \textsuperscript{1}University of Sheffield,
 \textsuperscript{2}AFP Medialab,
\\
 \small{
   \textbf{Correspondence:} \href{mailto:rosanna.milner@sheffield.ac.uk}{rosanna.milner@sheffield.ac.uk}
 }
}

\newcommand{\toolname}{\textsc{Verification Assistant}\xspace}
\newcommand{\pluginname}{\textsc{Verification Plugin}\xspace}

\begin{document}
\maketitle
\begin{abstract}

Disinformation and false content produced by generative AI pose a significant challenge for journalists and fact-checkers who must rapidly verify digital media information. While there is an abundance of NLP models for detecting credibility signals such as persuasion techniques, subjectivity, or machine-generated text, such methods often remain inaccessible to non-expert users and are not integrated into their daily workflows as a unified framework. This paper demonstrates the \toolname, a browser-based tool designed to bridge this gap. The \toolname, a core component of the widely adopted \pluginname (140,000+ users), allows users to submit URLs or media files to a unified interface. It automatically extracts content and routes it to a suite of backend NLP classifiers, delivering actionable credibility signals, estimating AI-generated content, and providing other verification guidance in a clear, easy-to-digest format. This paper  showcases the tool's architecture, its integration of multiple NLP services, and its real-world application to detecting disinformation.


\end{abstract}

\section*{Acknowledgments}
This  work  has  been  co-funded  by  the  UK’s innovation agency (Innovate UK) grant 10039055 (approved under the Horizon Europe Programme as vera.ai,  EU  grant  agreement  101070093) under action number 2020-EU-IA-0282. 
\section{Introduction and Related Work}\label{sec:introduction}

Digital disinformation poses a significant threat to democratic societies. The rapid advancement of generative AI, which can produce plausible text, images, and videos in seconds \cite{zhou2023survey}, exacerbates this problem. This technological shift has dramatically increased the volume and sophistication of ``fake news'', making the manual verification of online content a near-impossible task for journalists and fact-checkers \cite{guo2022survey}.

In response, the NLP community has developed a broad range of automated methods for information verification \cite{sharma2019combating, srba2024surveyautomaticcredibilityassessment}. Beyond high-level fake news detection, recent works have shifted toward more fine-grained, explainable assessments of content credibility levels that mirror the aspects of professional fact-checking \cite{shu2017fakenewsdetectionsocial, srba2024surveyautomaticcredibilityassessment}. Some examples of such credibility signals are the presence of propaganda and persuasion techniques \cite{piskorski-etal-2023-semeval}, bias and subjectivity  \cite{maab-etal-2024-media, piskorski-etal-2023-semeval}, and AI-generated text \cite{gehrmann-etal-2019-gltr}.

However, a significant gap persists between this state-of-the-art research and its practical application by journalists. Most verification tools require technical expertise, and are published as stand-alone models \cite{srba2024surveyautomaticcredibilityassessment}. This leaves non-technical users without efficient access to the very tools designed to help them. Among the existing  systems to support journalistic work and assist general users, the majority focus on a single functionality, such as news bias detection by AllSides\footnote{https://www.allsides.com} and GroundNews\footnote{https://ground.news/extension}, claim verification \cite{arora-etal-2019-claimbuster},  or propaganda detection \cite{da2020prta}.

To bridge this ``research-to-practice'' gap, we introduce the \toolname, a tool designed to integrate multiple backend NLP and media analysis microservices, making state-of-the-art research accessible within the user's browser. It represents an NLP-focused component within the \pluginname, a Chrome extension with over 140,000 active users\footnote{http://u.afp.com/plugin}. The \toolname provides a single, unified interface where a user can submit a URL (from a news article or social media post) or upload their own media. The system first extracts relevant content (text, metadata, images) and then dispatches this content to a configurable set of backend NLP microservices (e.g., machine generated text detection, topic and genre classification, URL domain analysis).
The \toolname's feature set is the result of continuous, participatory design sessions with a panel of journalists and fact-checkers, ensuring its real-world utility. A key focus of our design lies in supporting the multilingual nature of journalism.
Finally, the tool aggregates and displays the results in an informative, user-friendly dashboard, moving beyond simple binary ``fake/real'' labels and providing  nuanced, explainable insights.

\section{Architecture}\label{sec:archetecture}

The \toolname is a verification ``Swiss army knife'' that provides a single entry point to various state-of-the-art verification models.
There are three main components: (a) the frontend interface, (b) the assistant backend, and (c) the verification services.
The frontend provides the main user interface and is written in JavaScript using React Redux. 
It is distributed as a Chrome extension with language support for English, French, Spanish, Greek, Italian, Arabic, German, Japanese, Portuguese and Hungarian.
The code is open source under an MIT licence and is available at \url{https://github.com/AFP-Medialab/verification-plugin}.
The assistant backend is a Quart server application written in Python, with the source code stored in a private repository hosted by the GATE\footnote{\url{https://gate.ac.uk/}} team at the University of Sheffield.
It mediates interaction between the frontend and the verification services. Finally, the verification services host individual NLP models trained to assess specific credibility indicators, such as  presence of framing in the input text or the likelihood that the content is machine-generated. 

\begin{figure*}[ht!]
    \centering
    \resizebox{\textwidth}{!}{\input{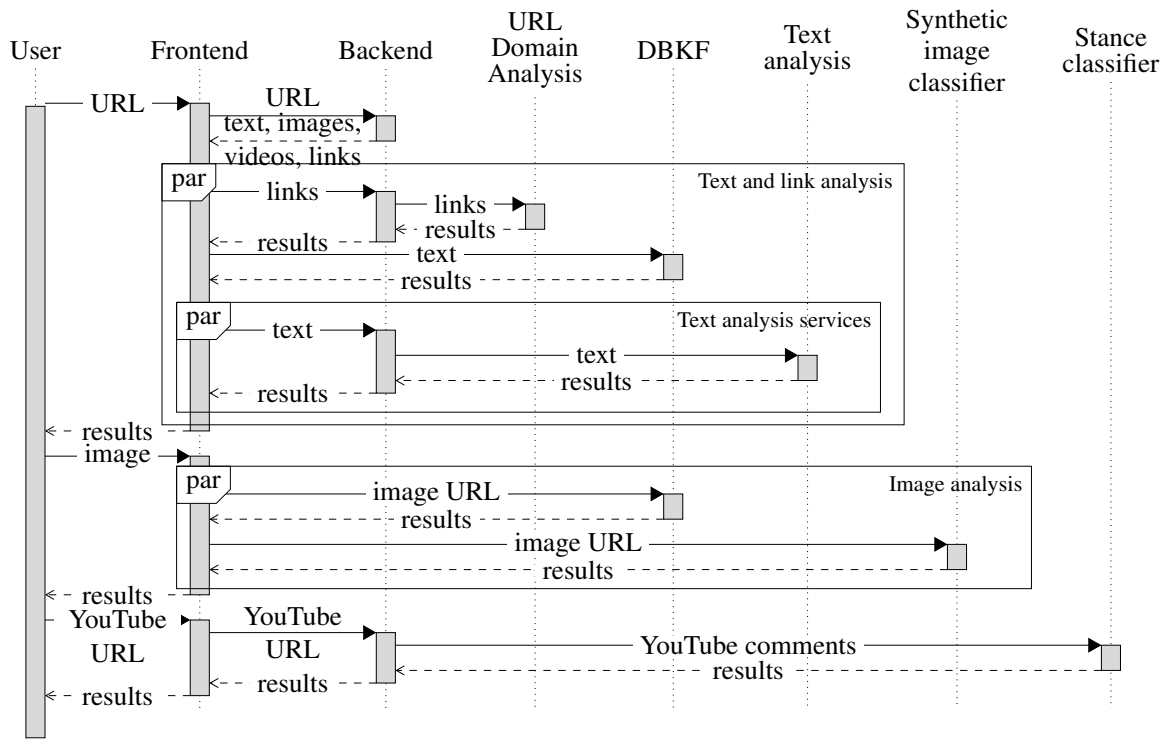}}
    \caption{Requests sent by the assistant when checking a typical webpage.}
    \label{fig:requests}
\end{figure*}

The main workflow is illustrated in \Cref{fig:requests}.
To verify an article, post, or item of media, the user can submit a URL through the frontend interface.
Alternatively, there is an option to upload an image or video file stored locally.
The frontend then passes this through to the assistant backend, which scrapes the contents and extracts the text, images, videos, and links contained in it and returns these to the frontend.
When the frontend receives the scraped results, it then sends several requests in parallel to the various verification services.
The remainder of this section introduces these services and provides details of their results in the order they appear in the frontend interface.

\subsection{Database of Known Fakes}
\label{sec:dbkf}
\begin{figure}[ht!]
    \centering
        \includegraphics[width=\columnwidth,trim={14cm 15cm 8cm 26cm},clip]{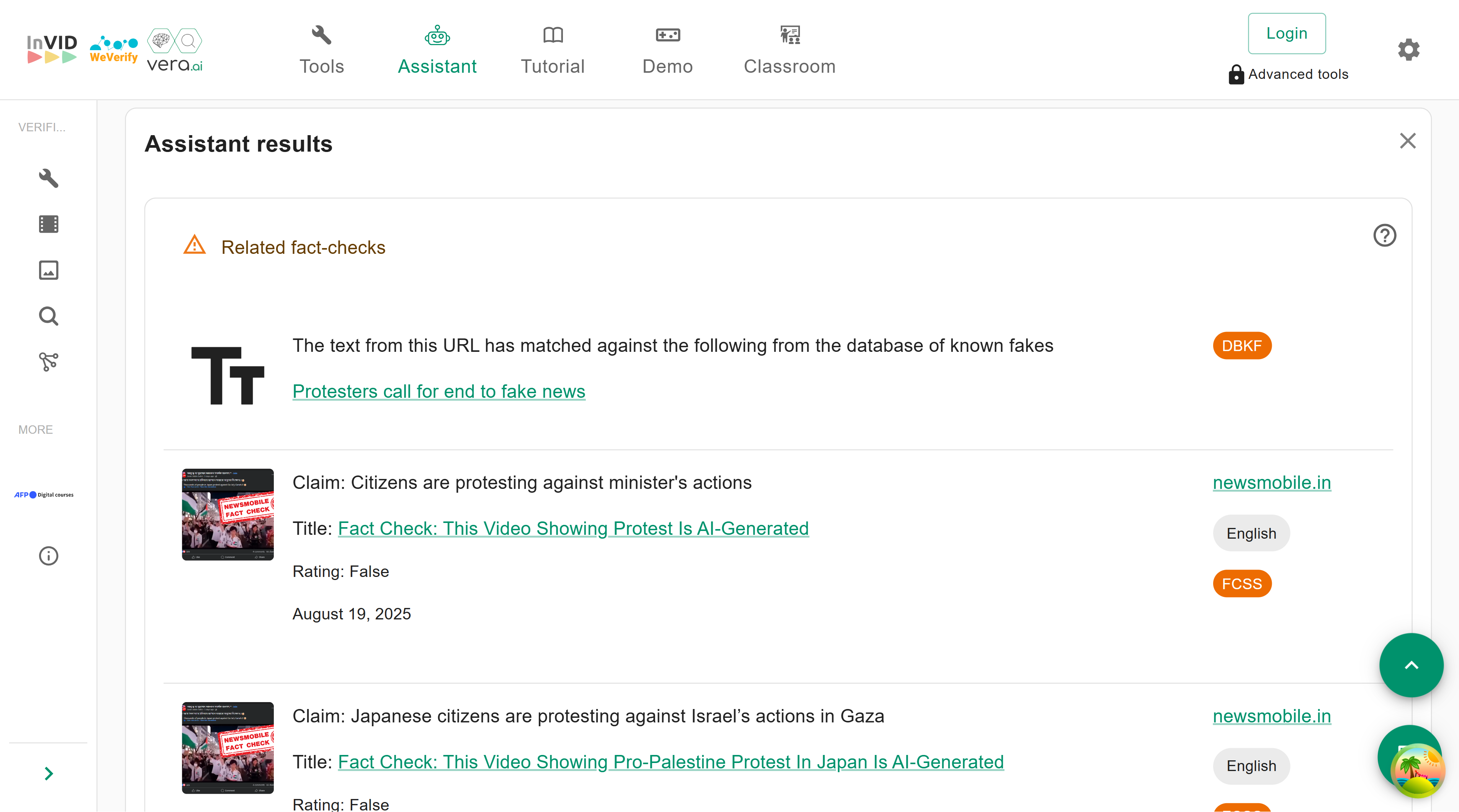}
    \caption{DBKF text service and Fact Check Semantic Search for simulated data.}
    \label{fig:dbkf}
\end{figure}

The Database of Known Fakes (DBKF)\footnote{\url{https://www.ontotext.com/knowledgehub/current/weverify-project/}} stores the claims that have been previously debunked by trusted organisations, such as Snopes\footnote{\url{https://www.snopes.com/}}. Its multilingual text service passes the extracted text through this database to find a potential match. It considers the first 100 characters from the text to imitate a title and prevent distant matches from occurring. 
If any matches are found, these are returned to the frontend along with a score for each match which are used for ranking purposes. For example, if one result has a higher score than a second result, then that first result is more relevant for the search.
Matches with a score ranking over 40 are presented to the user along with a link to the original debunk in the \textit{Detection of previously fact checked claims} section, as shown in \Cref{fig:dbkf}.

The Fact Check Semantic  (FCSS)\footnote{\url{https://kinit.sk/}} is the second part of the component  that matches a larger proportion of the text to other pieces of text that already exist in a collection of fact-checking databases, such as Snopes. 

\subsection{URL Domain Analysis}
\begin{figure}[!ht]
    \centering
    \begin{subfigure}{\columnwidth}
        \centering
        \includegraphics[width=\columnwidth,trim={36cm 39cm 4.7cm 10.9cm},clip]{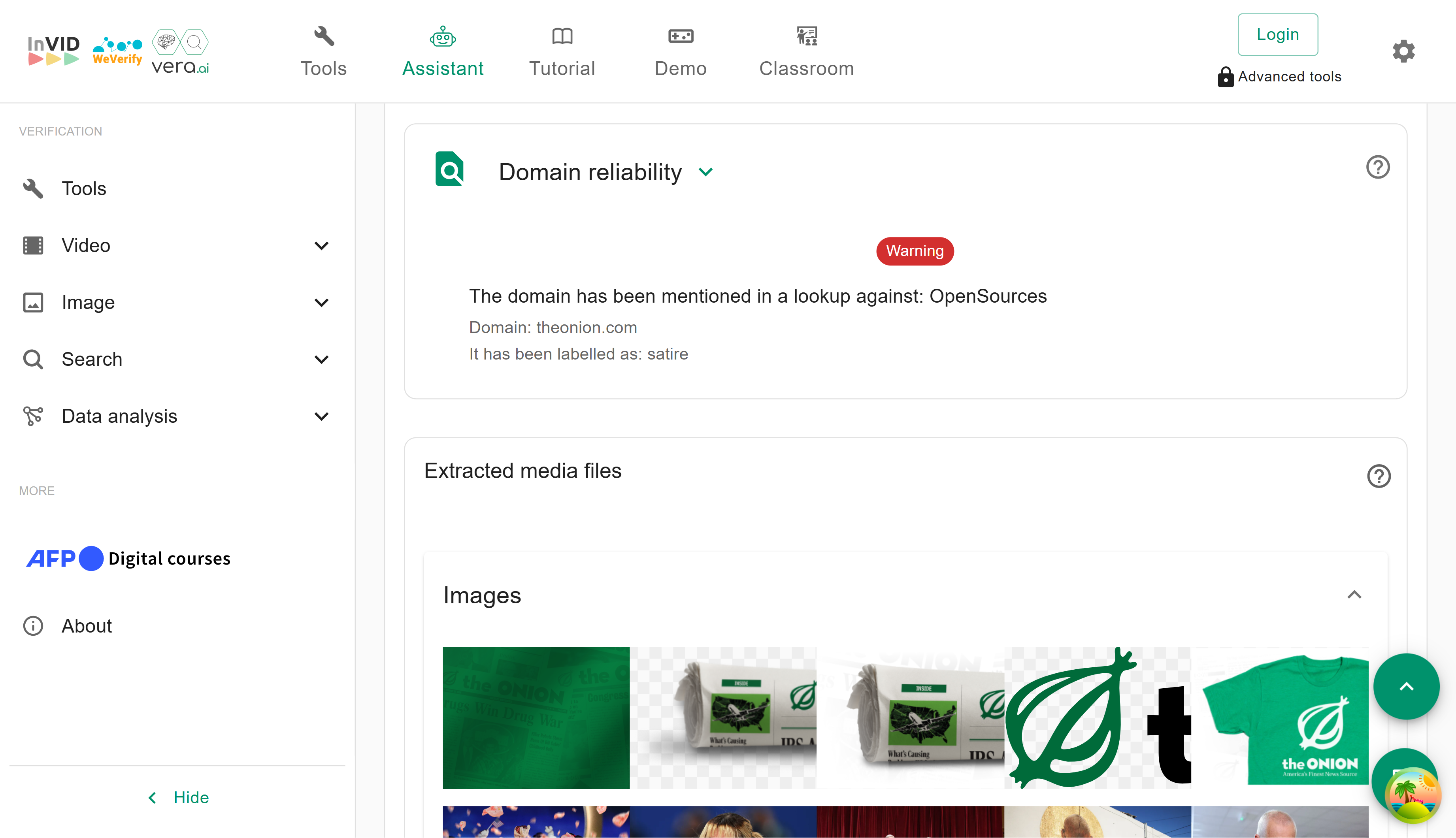}
        \caption{URL domain analysis results for submitted URL.}
        \label{fig:submitted-domain-analysis}
    \end{subfigure}
    \begin{subfigure}{\columnwidth}
        \centering
        \includegraphics[width=\columnwidth,trim={28cm 5cm 19cm 45cm},clip]{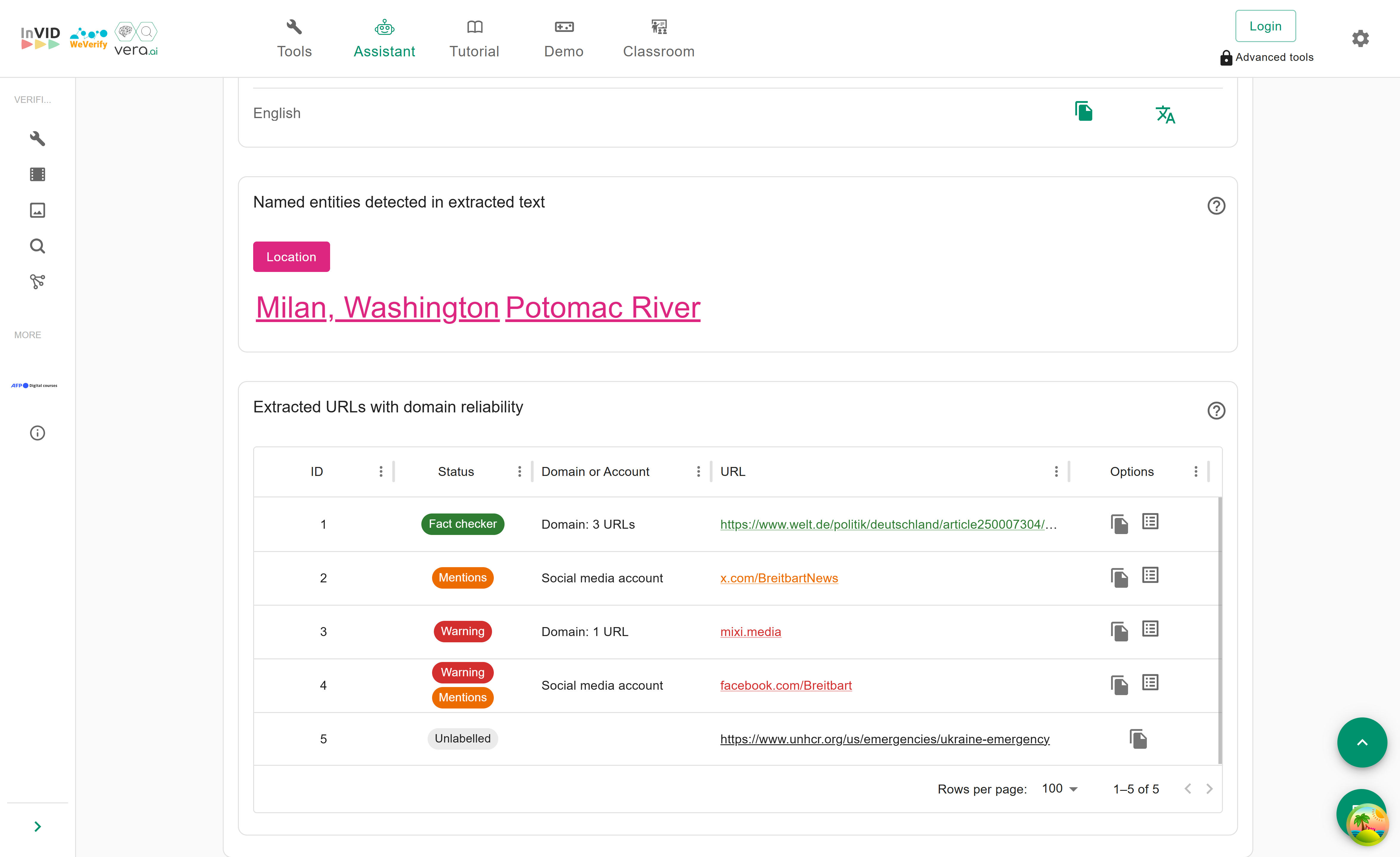}
        \caption{List of extracted URLs with URL domain analysis results using mock data.}
        \label{fig:link-domain-analysis}
    \end{subfigure}
    \caption{URL domain analysis results.}
    \label{fig:domain-analysis}
\end{figure}

The URL Domain Analysis service\footnote{\url{https://cloud.gate.ac.uk/shopfront/displayItem/url-domain-analysis}} collects information about a domain from multiple sources to inform the user about its credibility.
For example, the Duke Reporter's Lab\footnote{\url{https://reporterslab.org/}} maintains a database of known fact-checking sites.
Analysis of social media links is performed on individual accounts rather than site domains.
For example, if \url{https://x.com/BBCNews} was referenced in an article, the service would consider the BBCNews account rather than the entire ``x.com'' domain.
The URL for a site that has been explicitly listed as unreliable is flagged to the user as a \emph{Warning}, as shown in \Cref{fig:submitted-domain-analysis}.
Where the site has been mentioned as part of a debunk, but is not listed as unreliable, this is flagged to the user as a \emph{Mention} together with the details.
Known fact-checking sites are flagged to the user as a \emph{Fact Checker}.

Extracted links are additionally passed to the source credibility service, with warnings, mentions, and known fact-checking sites being flagged to the user.
The results are then grouped into domains and social media accounts, and are presented in a sortable grid format in the \textit{Extracted URLs with URL Domain Analysis} section, as illustrated in \Cref{fig:link-domain-analysis}.



\subsection{Media Analysis}
\begin{figure}
    \centering
    \includegraphics[width=\linewidth]{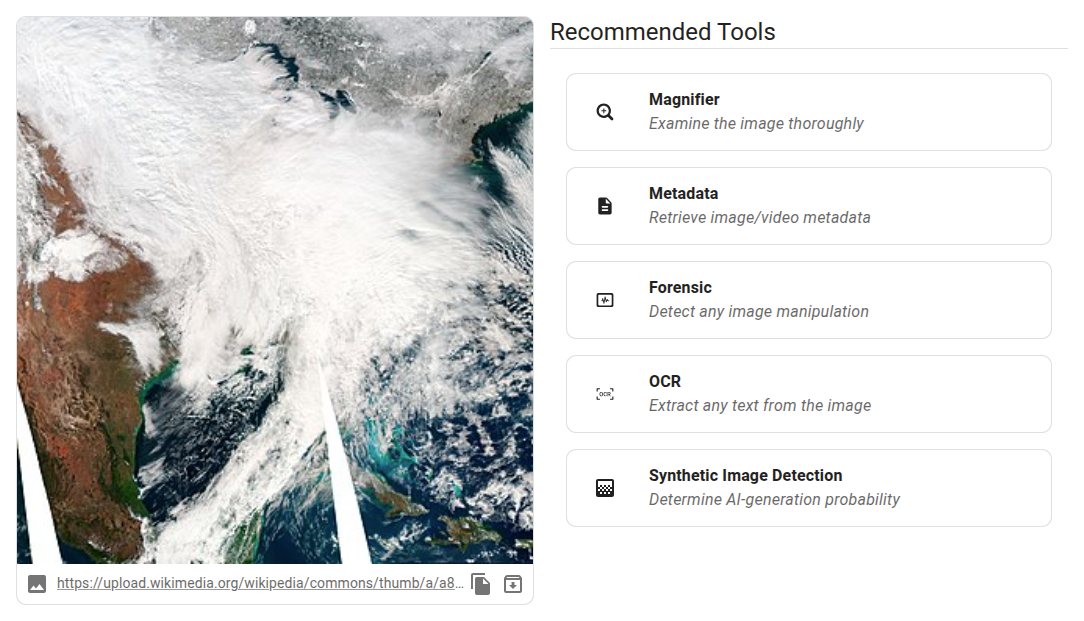}
    \caption{Recommended tools for an image.}
    \label{fig:recommended-tools-image}
\end{figure}

The \toolname displays extracted image and video thumbnails in the \textit{Extracted media files} section in the order in which they appear on the webpage, to facilitate an easy localisation of media in its original context.
illustrated in \Cref{fig:recommended-tools-image}, users may click on an extracted image or video to view more details about it, including a list of tools within the \pluginname that are recommended for analysing the media.
%
%
%
The possible recommended image analysis tools include image magnifier, metadata retrieval, forensic analysis, OCR, synthetic image detection, geolocalizer and provenance (C2PA). The video analysis tools include video analysis, keyframes, thumbnails, metadata and deepfake. The corresponding image and video analysis tools have been developed by AFP Medialab\footnote{\url{https://www.afp.com/en/fact-checking/fact-checking-afp/medialab}}, Borelli Center\footnote{\url{https://ens-paris-saclay.fr/en/research/research-laboratories/centre-borelli}}, ITI CERTH\footnote{\url{https://caa.iti.gr/}}, GRIP UNINA\footnote{\url{https://www.grip.unina.it/}} and the University of Sheffield\footnote{\url{https://cloud.gate.ac.uk/}}. 


Images are additionally sent to the DBKF image search service.
Using the image similarity techniques, the service searches for matches in a database of already debunked fakes.
If a match is found, this is highlighted to the user as a warning and displayed in the \textit{Detection of previously fact checked claims} section, as described in \Cref{sec:dbkf}.

\subsection{Credibility Signals}\label{sec:credibility-signals}

The term \textit{credibility signals} refers to a set of context- and content-based indicators that cumulatively contribute to the overall assessment of the credibility of textual information \cite{srba2024surveyautomaticcredibilityassessment}.
The \toolname incorporates 5 such content-based signals proven to be vital for credibility assessment by prior research \cite{srba2024surveyautomaticcredibilityassessment}. More specifically, it integrates state-of-the-art classifiers for Framing \cite{razuvayevskaya2024comparison, wu2023sheffieldveraai}, Genre \cite{razuvayevskaya2024comparison, wu2023sheffieldveraai}, Persuasion Techniques \cite{razuvayevskaya2024comparison, wu2023sheffieldveraai}, Subjectivity \cite{schlicht2023dwreco}
and Machine Generated text \cite{macko-etal-2023-multitude} detection. Each signal detector is implemented through a classifier that outputs the associated signal label along with the location in text where the signal is most likely to be present.
Together, these credibility signals can be interpreted by the user as an ``information nutrition label'' \cite{fuhr2018information}. The user can find the results for the credibility signals in the \textit{Extracted text} section separated into different tabs.

\begin{figure}[t!]
    \centering
    \begin{subfigure}{\columnwidth}
        \centering
        \includegraphics[width=\linewidth,trim={0cm 8.5cm 0cm 0cm},clip]{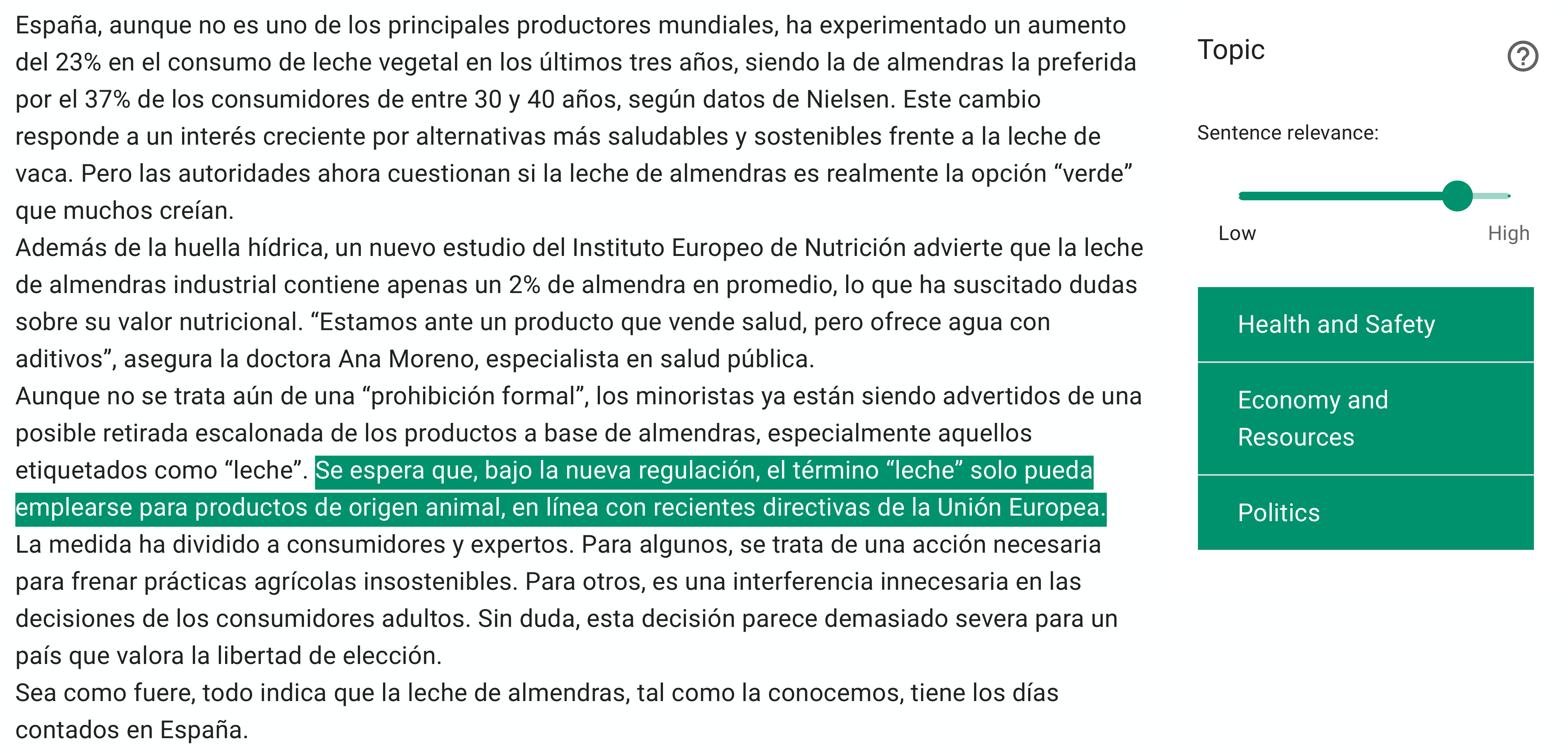}
        \caption{News topic service.}
        \label{fig:credibility-signals-news-framing}
    \end{subfigure}
    \begin{subfigure}{\columnwidth}
        \centering
        \includegraphics[width=\linewidth]{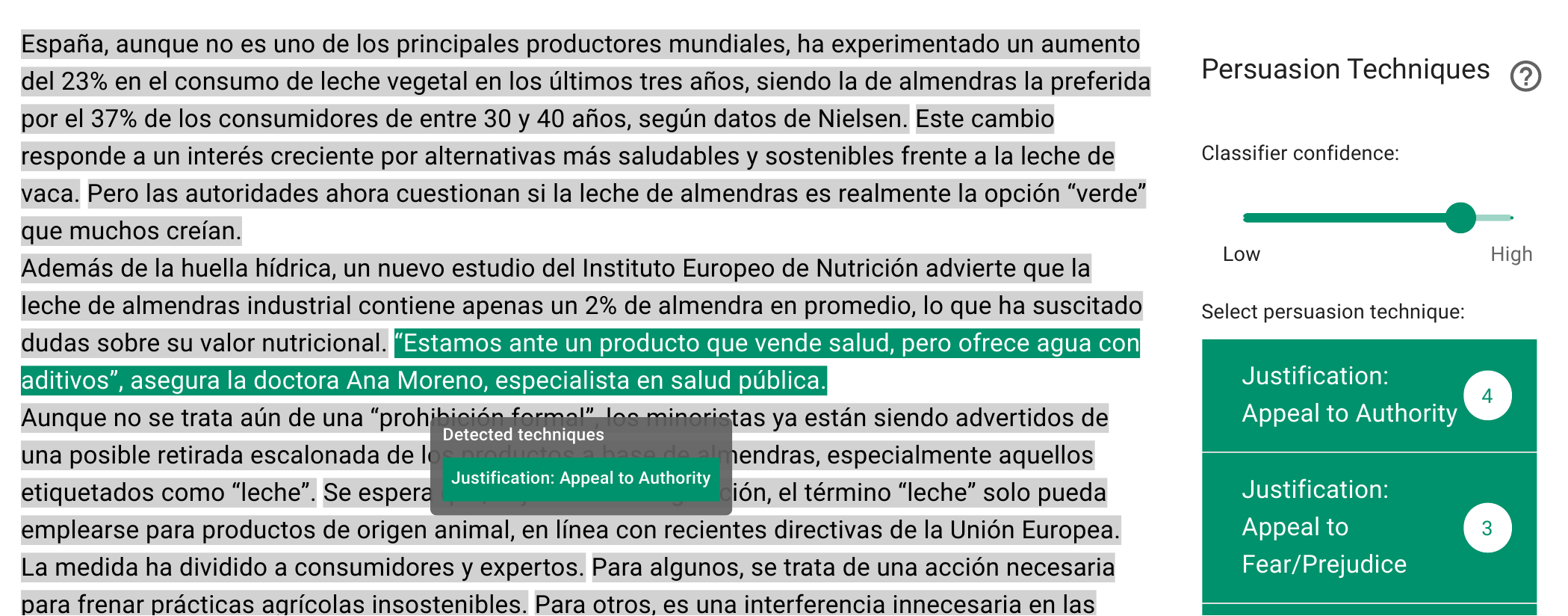}
        \caption{Persuasion techniques service.}
        \label{fig:credibility-signals-persuasion-techniques}
    \end{subfigure}
    \begin{subfigure}{\columnwidth}
        \centering
        \includegraphics[width=\linewidth,trim={0cm 3.5cm 0cm 0cm},clip]{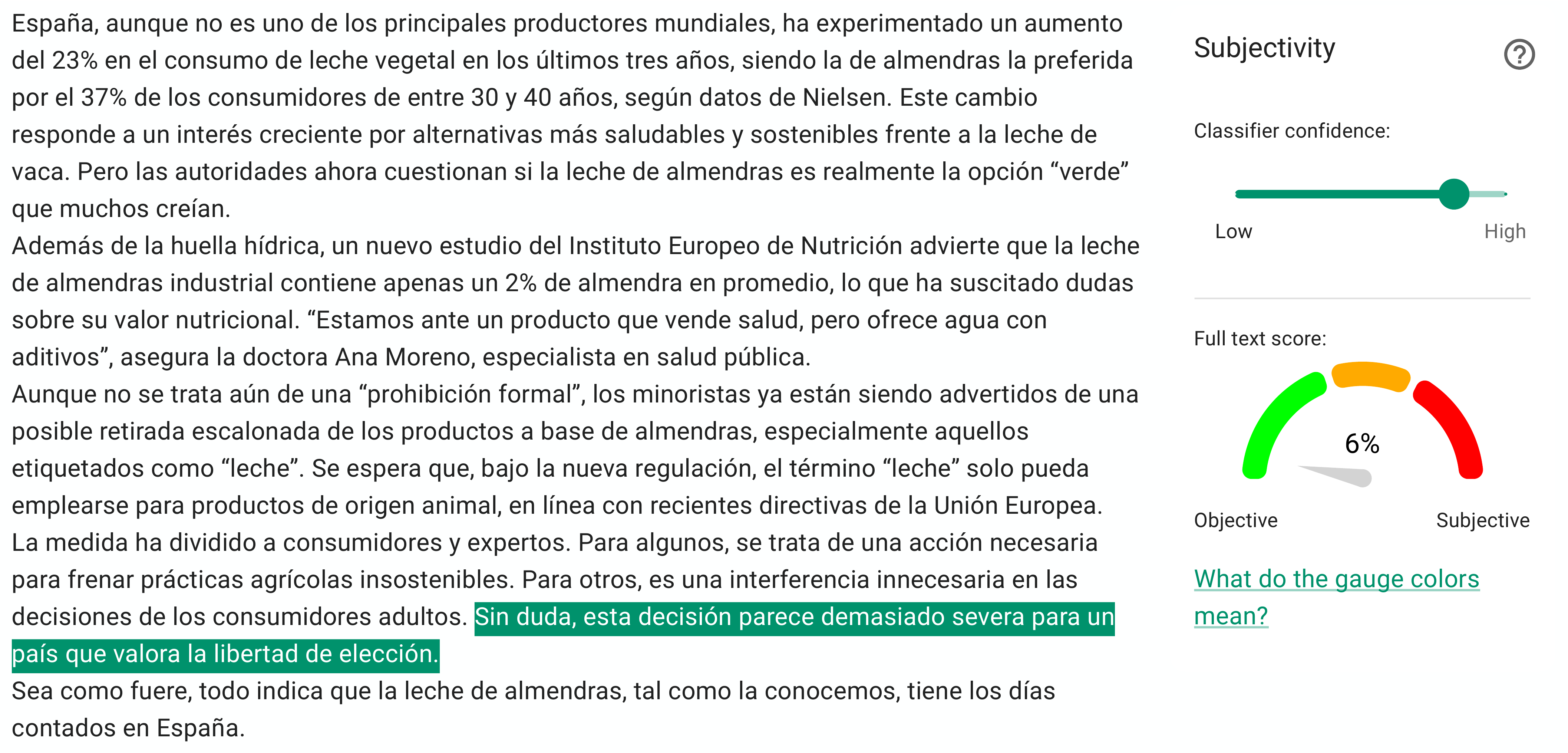}
        \caption{Subjectivity service.}
        \label{fig:credibility-signals-subjectivity}
    \end{subfigure}
    \begin{subfigure}{\columnwidth}
        \centering
        \includegraphics[width=\linewidth]{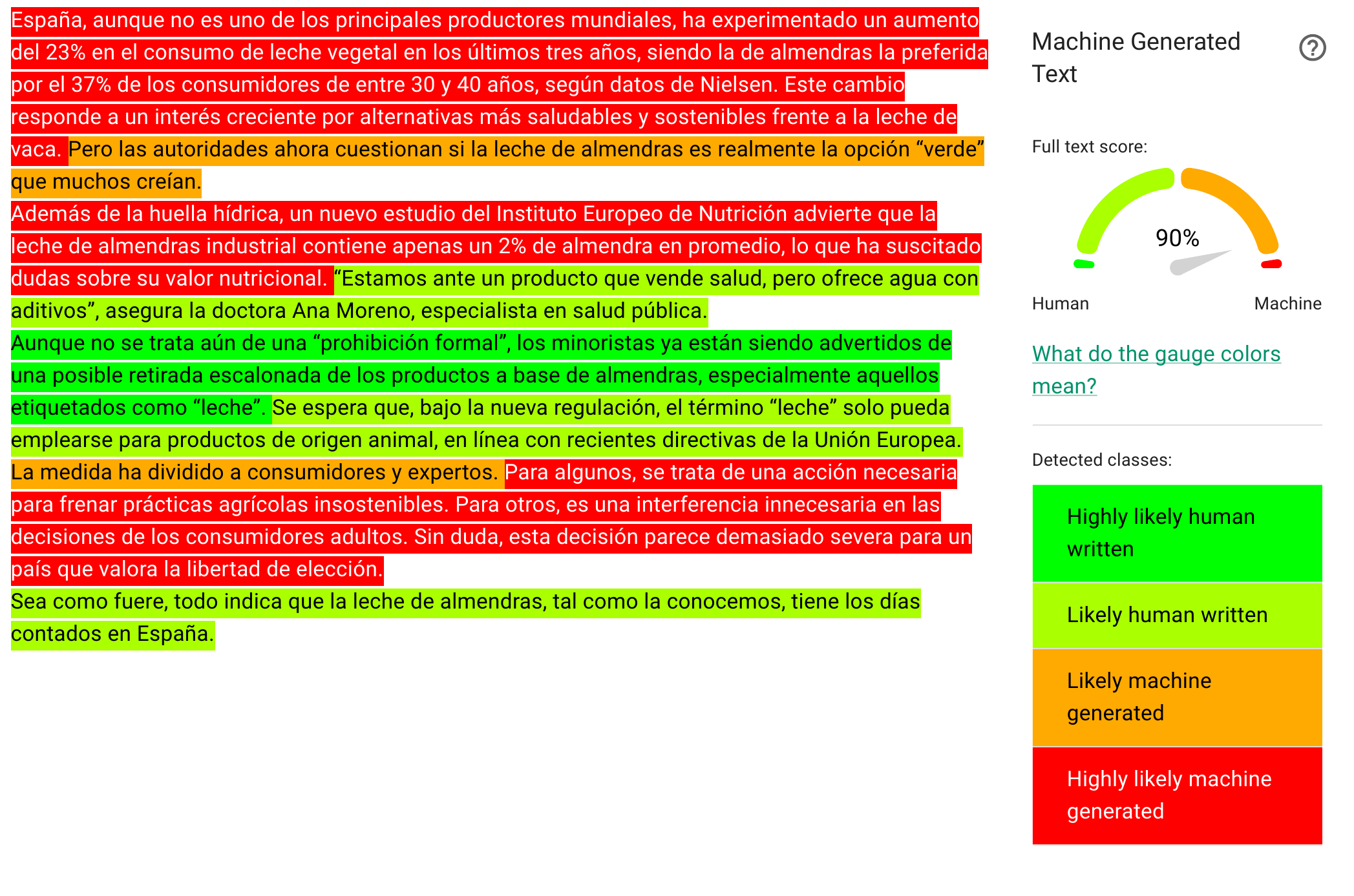}
        \caption{Machine generated text service.}
        \label{fig:credibility-signals-machine-generated-text}
    \end{subfigure}
    \caption{Credibility signals on an AI-generated article.}
    \label{fig:credibility-signals}
\end{figure}

\subsubsection{Framing} \label{sec:framing}

Framing, referred in \toolname as \textit{Topic}, is a signal that represents the perspective from which information is presented \cite{srba2024surveyautomaticcredibilityassessment}. Its purpose is to “frame” information and guide readers toward a particular meaning. The framing classifier \cite{razuvayevskaya2024comparison, wu2023sheffieldveraai} has been trained to detect nine main frames: \textit{Economy and Resources}, \textit{Religious, Ethical and Cultural}, \textit{Fairness, Equality and Rights}, \textit{Law and Justice System}, \textit{Crime and Punishment}, \textit{Security, Defense and Well-being}, \textit{Health and Safety}, \textit{Politics and International Relations}. A single news article frequently incorporates several overlapping frames simultaneously, and the tool returns all the frames above a certain threshold. For example, \Cref{fig:credibility-signals-news-framing} shows the top three that have a confidence score $>0.8$. The classifier has been tested on six languages used during fine-tuning and three languages ``unseen" during training \cite{razuvayevskaya2024comparison}. The model demonstrated average performance of $F_{1_{macro}}=59.9\pm3.1$ and $F_{1_{micro}}=61.7\pm7.5$ across all 9 languages.  
The interface highlights sentences that are deemed important by the underlying model in making a decision based on the normalised per-sentence attention scores. The interface allows the user to change the sentence importance threshold by moving a slider. 

\subsubsection{Genre}

Information \textit{genre} is often defined as a way of distinguishing texts based on their writing style \cite{srba2024surveyautomaticcredibilityassessment}. For the purpose of information credibility detection, the distinction most relevant to the task concerns whether a text maintains an objective tone or incorporates manipulative language. To operationalize this, the incorporated classifier \cite{razuvayevskaya2024comparison, wu2023sheffieldveraai} distinguishes between three types of genre: \textit{objective} reporting, \textit{opinionated} pieces, and \textit{satirical} content. Within this taxonomy, objective reporting is characterized by comprehensive coverage of pertinent facts and perspectives. Opinionated news, on the other hand, tends to rely on persuasive or propagandistic techniques, while satirical articles differ from both categories, as they intentionally employ fictionalized material for comedic or critical purposes. Similarly to the framing classifier, the models were tested on both ``seen" and ``unseen" languages \cite{razuvayevskaya2024comparison}, with the overall  performance of $F_{1_{macro}}=49.2\pm7.4$ and $F_{1_{micro}}=56.7\pm6.1$, averaged across genres and languages.
The UI is  identical to the framing classifier described in Section~\ref{sec:framing}.


\subsubsection{Persuasion Techniques}

\textit{Persuasion techniques}, sometimes referred to as \textit{propaganda techniques} \cite{piskorski-etal-2023-semeval}, refer to  communication strategies aimed at influencing or manipulating the reader's opinions. The classifier \cite{razuvayevskaya2024comparison, wu2023sheffieldveraai} is trained to identify 23 different techniques which can be organised into six groups: \textit{justification}, \textit{simplification}, \textit{distraction}, \textit{call}, \textit{manipulative wording} and \textit{attack on reputation}. The model achieved an average performance of $F_{1_{macro}}=23.7\pm5.0$ and $F_{1_{micro}}=41.8\pm8.6$ across all persuasion techniques and 9 languages, 6 seen and 3 unseen.  

\Cref{fig:credibility-signals-persuasion-techniques} presents the user interface. Sentences with a detected persuasion technique(s) are shown as highlighted. The interface also allows the user to hover over a certain highlighted sentence and see the corresponding detected technique(s). A list of the persuasion techniques detected across the complete text is shown on the right hand side. The user can select any technique to only highlight the sentences in which it appears. For this credibility signal, the algorithm provides a confidence score for each sentence-persuasion technique pair. Similarly to the framing classifier,  a slider can be moved by the user to change the threshold of these confidence scores for highlighting the associated sentences. 


\subsubsection{Subjectivity}

\textit{Subjectivity} signal refers to the degree to which a news article reflects personal opinions, biases, or emotions rather than strictly objective facts. The integrated classifier \cite{schlicht2023dwreco}, trained based on the CLEF-2023 CheckThat! data challenge \cite{10.1007/978-3-031-28241-6_59}, returns the score indicating the degree of subjectivity per sentence. 
The classifier was trained in a multilingual manner, with the test and training sets in English, Turkish and German. The model demonstrated relative robustness across the languages, with the performance of $F_1=0.87$, $F_1=0.78$ and $F_1=0.74$ for Turkish, English and German respectively. 

As shown in \Cref{fig:credibility-signals-subjectivity}, the interface highlights the subjective sentences identified in text. The overall subjectivity score of the whole text is calculated as a percentage  and displayed in a gauge format. The percentage falls into one of three levels: \textit{objective}, \textit{somewhat subjective} and \textit{highly subjective}.
Similar to  persuasion techniques, a confidence score is provided for sentence subjectivity level and a slider can be moved by the user to change its  threshold. 

\subsubsection{Machine Generated Text}

\textit{Machine-generated text} detection as a credibility signal should be distinguished from legitimate automatically generated content, such as machine translation or grammar correction. The classifier \cite{macko-etal-2023-multitude} integrated into the tool, therefore, focuses solely on machine-generated text intended for malicious purposes, such as spreading misinformation. 
The \toolname automatically identifies how likely a piece of text is to be machine generated (e.g. by a large language model), and presents this to the user as a percentage of the full text in a gauge format, as seen in Figure \ref{fig:credibility-signals-machine-generated-text}. The text is split into sentences each of which belongs to one of four categories: \textit{highly likely human}, \textit{likely human}, \textit{likely machine generated} and \textit{highly likely machine generated}. The MGT service represents a call to a set of detectors, each trained to identify the generation by a various combinations of language models. Each model was fine-tuned on the data in English, Spanish and Russian languages, and subsequently tested on related languages for each of the training ones: Dutch and German for English,
Czech and Ukrainian for Russian, Portuguese and
Catalan for Spanish. The evaluation results \cite{macko-etal-2023-multitude} demonstrated high accuracy, with the best-performing model demonstrating $F_{1_{macro}}=0.8480$ and weighted $F_1=0.9400$, averaged across the classes, generator models and languages.

\subsection{Named Entities}
\begin{figure}
    \centering
    \includegraphics[width=\linewidth]{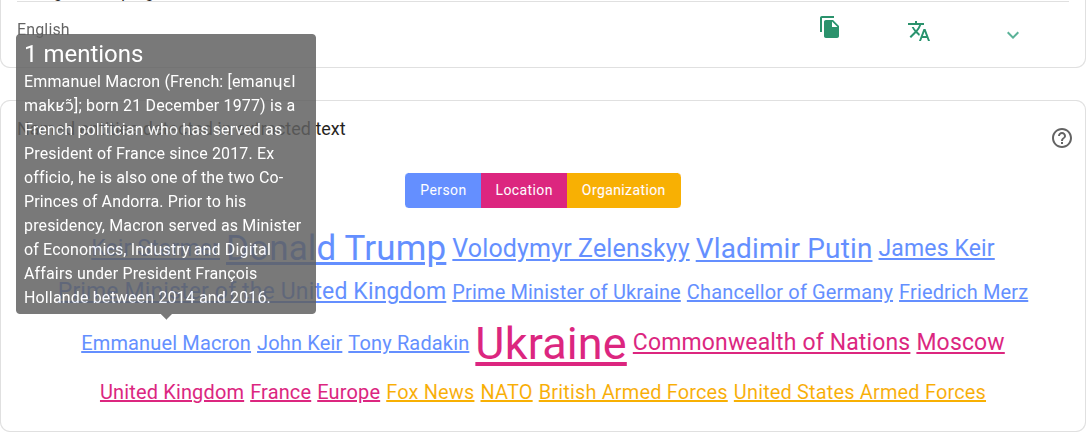}
    \caption{Named entities extracted from a news article on the war in Ukraine.}
    \label{fig:named-entities}
\end{figure}

The named entity detector provided by the University of Sheffield identifies the names of people, locations, and organisations in text documents.
To extract the named entities, the tool first identifies WikiData concepts in the text.
These are then linked to their corresponding DBpedia articles, from which the \texttt{rdf:type} field is used for classification.
Entities that are not classified as a \texttt{Person}, \texttt{Location}, or \texttt{Organization} are filtered out. The extracted entities are presented to the user as a word cloud, as illustrated in \Cref{fig:named-entities}. Here,  users can select which classes of entities are displayed.
The size of each entity in the cloud is proportional to the number of times that entity is mentioned in the text.
When the user hovers over an entity, they are shown an abstract from DBpedia along with how many times the entity was mentioned in the text and the link to the corresponding DBpedia article.

\subsection{YouTube Comments with Stance Classifier}
\label{sec:youtube}
\begin{figure}
    \centering
    \includegraphics[width=\columnwidth,trim={0.5cm 0 0cm 4cm},clip]{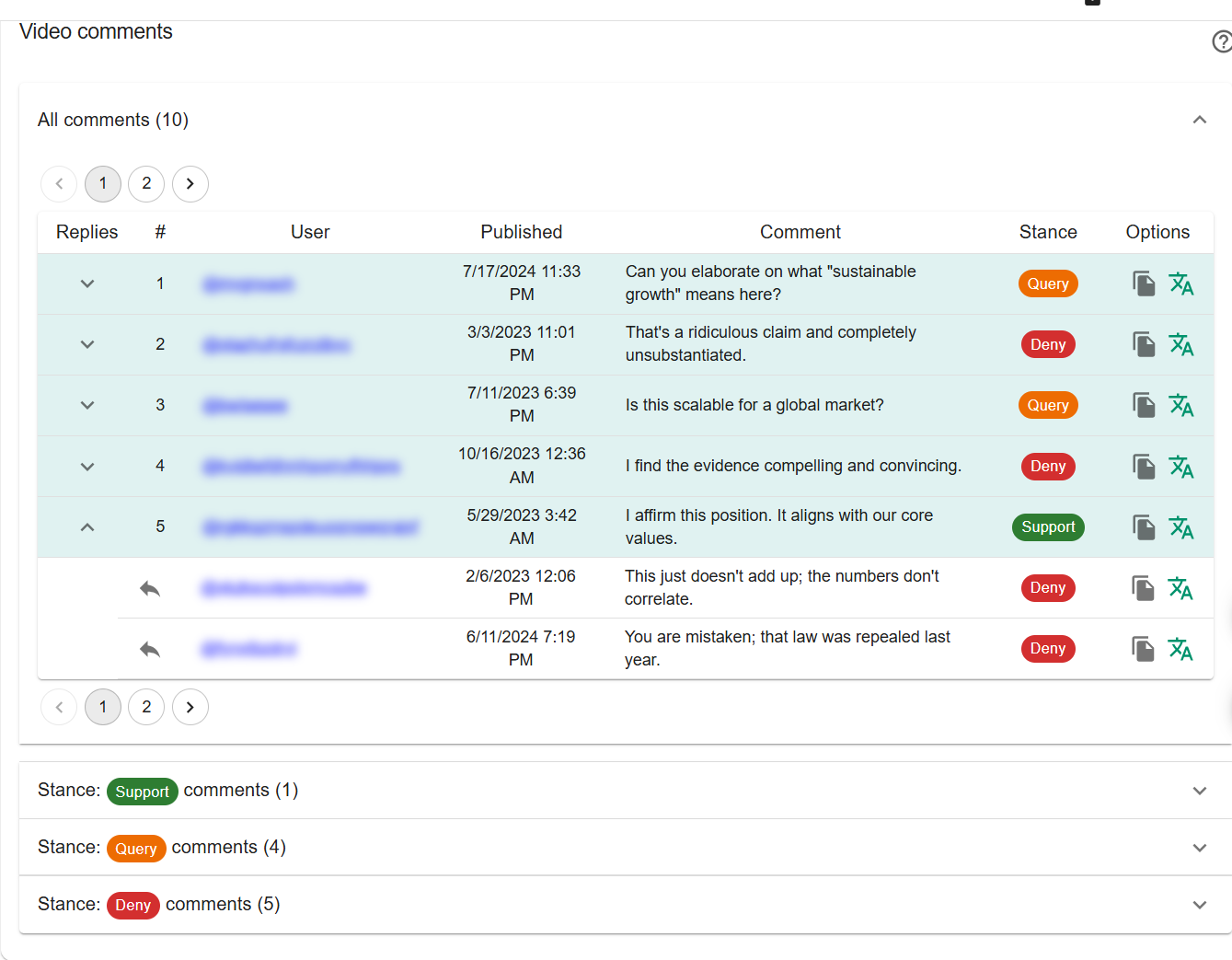}
    \caption{Synthesized comments with  classifications. Usernames have been blurred to avoid  accidental collisions with real (and potentially future) YouTube users.}
    \label{fig:youtube-comments}
\end{figure}

When the user submits the URL of a YouTube video, the backend calls the YouTube API to extract the top 10 video comments with their replies.
These comments and replies are sent to the stance classifier\footnote{\url{https://cloud.gate.ac.uk/shopfront/displayItem/stance-classification-multilingual}} which identifies comments that \textit{support}, \textit{deny}, or \textit{question} the video, with respect to its title.
For comment replies, the classifier identifies whether comments \textit{support}, \textit{deny}, or \textit{query} the original comment. If either are found to be none of these, then the comment is simply labelled as \textit{comment}.
This is illustrated in \Cref{fig:youtube-comments} which, to protect the privacy of real users, shows a set of synthesized comments from synthesized users, see \Cref{sec:appendix}.

\section{Evaluation}\label{sec:evaluation}
\begin{figure*}
    \centering
    \includegraphics[width=0.8\linewidth]{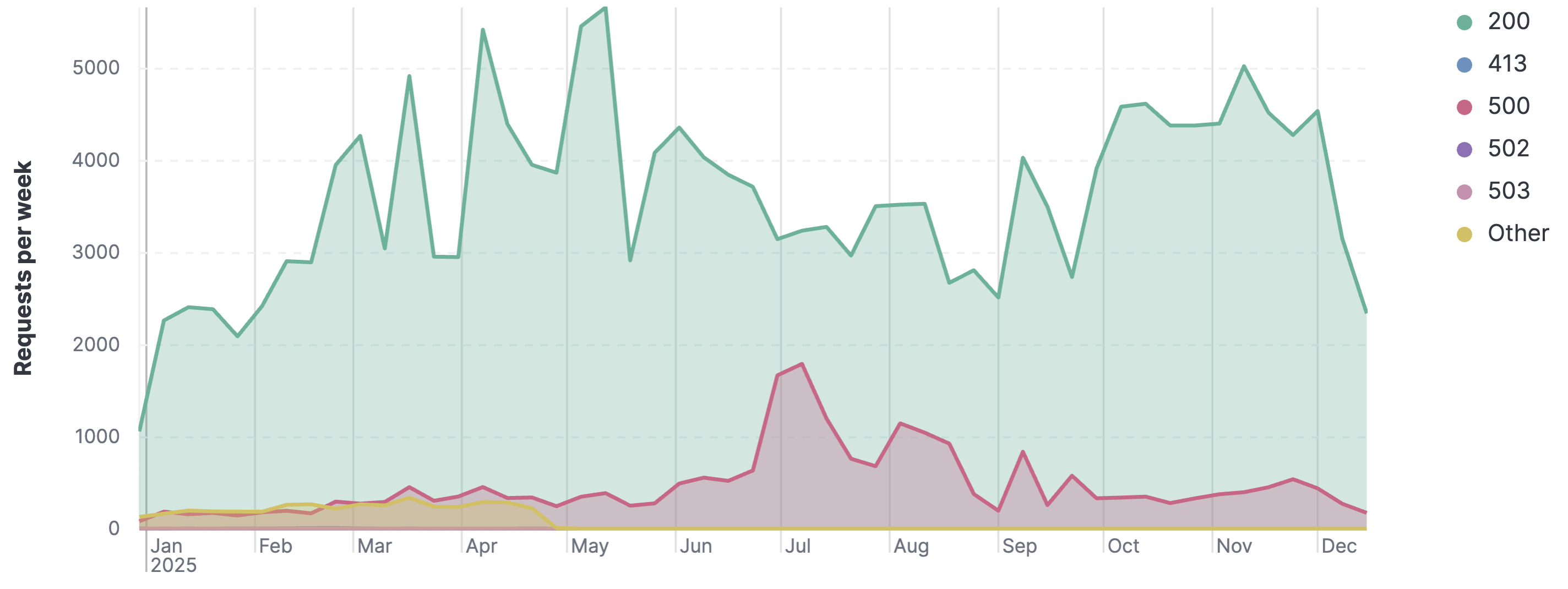}
    \caption{HTTP response code distribution through 2025.}
    \label{fig:responses}
\end{figure*}

We collected quantitative feedback from 72 target users--participants of IFCN Global Fact 12\footnote{Global Fact is the main yearly worldwide gathering of fact-checkers} conference--in the form of the questionnaires. The survey consisted of 16 questions about the participants, their main occupation, their frequent use of
the tool, the features’
usefulness in their workflow, and more qualitative questions about the user interface, their trust in the
results or new features they would like to see in updates. Respondents (mainly from AFP, Al Jazeera, dpa, LeadStories, MythDetector, Deutsche Welle and France24) could declare several occupations. Their frequency of use of the tool measured 3.75 out of 5, with 19.44\% of respondents using it very often and 41.67\% often. Users were asked to rate each new (or enhanced) feature in a scale between Excellent-Good-Average-Poor-Very poor, and this was later transformed into a Likert scale.  The overall  satisfaction with the \toolname scored 3.56 out of 5.

Additionally, regular user-centred design (UCD) evaluation sessions were conducted with our target users: professional fact-checkers, journalists, and disinformation researchers. 
During each session, participants first received a brief overview of the tool and its capabilities. They were then asked to verify a provided piece of information. In addition, participants were encouraged to submit their own content relevant to their daily work—for example, a news article in Greek—and discuss whether they considered the information trustworthy and described how the tool’s outputs influenced their judgment. The primary goal of these sessions was to assess the tool's practical utility and usability, rather than just raw classifier accuracy. Qualitative feedback was gathered on result agreement, clarity of the output, and the tool's overall value in their workflow. 
This feedback was central to our iterative development cycle, leading directly to practical UI refinements. Participants also suggested features or interface improvements that could enhance usability. For example, user feedback prompted the addition of explanatory pop-ups that clarify what each classifier measures and how to interpret its results. 

Alongside continuously gathering user feedback, we regularly collect logs of service calls and returned errors to evaluate the robustness of the \toolname. Figure~\ref{fig:responses} shows the distribution of HTTP response status codes from the start until the end of 2025. As can be seen, only 14.50\% of the requests received a server error response, with the majority of these errors being due to the scraping failures. Overall, 30.82\% of user requests resulted in scraper errors, which could be a result of the users submitting URLs that are either incorrect or cannot be accessed by the tool.  Overall, Figure~\ref{fig:asisst_calls} (Appendix~\ref{app:1}) demonstrates the clear growing trend from the beginning until the end of 2025 in terms of the number of queries submitted by the end-users, with an overall of 18,106 requests submitted from the tool page.

\section{Conclusions and Future Work}\label{sec:conclusion}
This paper has presented the \toolname, a Chrome extension that directly bridges the gap between advanced NLP research and the daily workflow of journalists and fact checkers. By providing a common interface to a plethora of text classifiers, the \toolname empowers users
to analyse content in terms of AI-generation, subjectivity, and other credibility signals.
The \toolname's integration within a plugin used by over 140,000 users demonstrates its real-world value. Its design is continually refined based on feedback from a panel of fact-checkers, journalists and researchers, ensuring it remains relevant to user needs.

Future work is divided into two main streams. From an engineering perspective, the primary challenge is to develop robust, long-term solutions for content scraping from social media platforms and news sites, which frequently alter their structure and limit access. From a research perspective, we aim to enhance the \toolname's utility by integrating more complex AI models. Additionally, increasing the transparency and explainability of their outputs is paramount for building and maintaining user trust. Finally, the recent migration from Webpack to WXT has enabled browser-agnostic functionality, with Firefox support currently under development.





\bibliography{bibliography}

\appendix

\section{User request trends} \label{app:1}
\begin{figure*}[ht!]
    \includegraphics[width=1\linewidth]{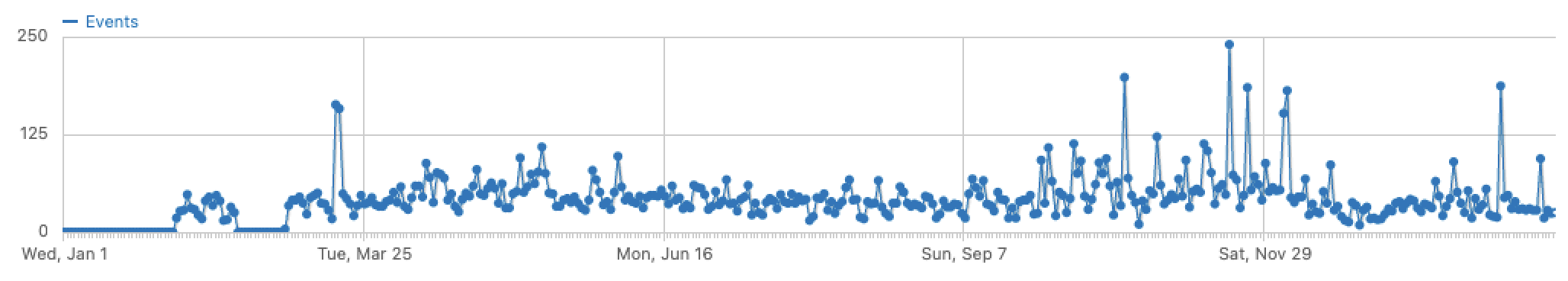}
    \caption{Temporal distribution in the number of events (queries) submitted from the \toolname page during 2025.}
    \label{fig:asisst_calls}
\end{figure*}
Figure~\ref{fig:asisst_calls} shows the number of unique requests or inputs submitted by the users during 2025. 
A clear growing trend can be observed in terms of the increasing number of events. 
\section{Prompt for synthesised YouTube comments}\label{sec:appendix}
The prompt given to Gemini to generate mock YouTube comments and replies for \Cref{fig:youtube-comments}: ``I am trying to test a stance classifier. The tool takes a string representing a YouTube comment or twitter post or something and classifies it as either supporting, denying, or questioning a particular statement, or just as a `comment' if it doesn't support, deny, or question. I need some sample test input for this tool that doesn't come from anywhere. Please can you generate 100 random inputs with a selection of support, deny, questions and comments. With one input on each line?''

\end{document}